# Predicting Fruit Quality with a Hybrid Machine Learning and Image Processing Approach

Amir Reza Hashemi[1]*, Shahram Amiri[2]

[1] Department of Computer Science, Science and Research Branch, Islamic Azad University, Tehran, Iran

[2] Department of Decision and Information Sciences, School of Business Administration, Stetson University, DeLand, Florida, USA

* Corresponding author: amirrezahmi2002@gmail.com

**Abstract**

Fruit spoilage is a significant issue in agriculture, leading to substantial economic losses. Addressing this, our study introduces a hybrid approach combining image processing and deep learning to assess fruit freshness. We developed an image processing algorithm that quantifies spoilage on a scale from 0 (fully fresh) to 100 (fully rotten). Alongside, we trained a convolutional neural network (CNN) to perform binary classification (fresh or rotten) using a large dataset of fruit images. The outcomes of both methods were synthesized using logistic regression to enhance the accuracy of freshness predictions. Subsequently, this logistic regression model was utilized to enable the image processing algorithm to provide binary classification based on its percentage output, thus eliminating the need for the CNN in real-time applications. Our approach, which does not require high computational resources, achieved real-time performance and was validated with over 90% accuracy on a dataset comprising apples and oranges. The primary limitation lies in the requirement for fruits to be isolated on a background that must be either white or transparent, suggesting future improvements could include advanced segmentation models to automate background removal. This study's results highlight the potential of integrating simple image processing techniques with machine learning to provide practical solutions in the agricultural sector.

## 1 Introduction

The agriculture industry is a vital sector of the global economy, providing food and other products for consumption (Olaoye, 2014). However, one major challenge facing the industry is the problem of fruit spoilage (Atanda et al., 2011). Fruit spoilage not only results in significant economic losses but also poses a threat to food security (Rawat, 2015). Various methods have been proposed to tackle this issue, ranging from traditional visual inspection to advanced technological solutions, including spectral analysis and chemical sensors.

In this study, we propose a hybrid approach for predicting the freshness of fruit using both image processing and machine learning techniques. Unlike existing methods, which often rely on either detailed chemical analysis or require sophisticated hardware, our approach combines a simple image processing algorithm with a convolutional neural network (CNN) model to classify fruit as either fresh or rotten based on visual features. This novel integration allows for a quick assessment of fruit condition without the need for extensive preprocessing or complex setups.

The proposed approach begins by collecting and preprocessing a dataset of images of fruit, consisting of both fresh and rotten examples. Using this dataset, we train a convolutional neural network (CNN) model to classify the fruit as either fresh or rotten. We also develop an image processing algorithm that estimates the percentage of spoilage in the fruit based on visual features. We then use logistic regression to combine the output of the CNN model with the output of the image processing algorithm, enhancing the accuracy of predicting the freshness of the fruit. Subsequently, the logistic regression model was utilized to enable the image processing algorithm to provide binary classification based on its percentage output. This eliminates the need for the CNN model in real-time applications, making our method both efficient and resource-friendly.

The performance of our hybrid approach is evaluated on a validation set, and the results demonstrate its effectiveness at accurately predicting the freshness of the fruit. The proposed approach has the potential to be applied in real-world settings, such as in the agricultural industry, to improve the efficiency and effectiveness of fruit inspection and grading.

In the following sections, we will discuss the methods used to collect and preprocess the dataset, the implementation of the CNN and image processing algorithms, the results obtained from the evaluation of the proposed approach, and its potential applications and future work.

## 2 Data Collection and Preprocessing

The dataset used in this study was obtained from Kaggle and includes images of fresh and rotten apples and oranges (Kalluri, 2018). The dataset was divided into two main categories: "fresh" and "rotten". Each of these categories contains images of both apples and oranges. The dataset was divided into a training set and a validation set, with the training set being used to train the CNN model and the validation set being used to evaluate the performance of the proposed approach.

The training and testing datasets comprise a significant number of images across different fruit classes. The training dataset includes:

- 1,693 files of fresh apples,
- 1,466 files of fresh oranges,
- 2,342 files of rotten apples,
- 1,595 files of rotten oranges.

The testing dataset includes:

- 395 files of fresh apples,
- 388 files of fresh oranges,
- 601 files of rotten apples,
- 403 files of rotten oranges.

**Note:** The dataset also contained images of bananas, which were not used in this study.

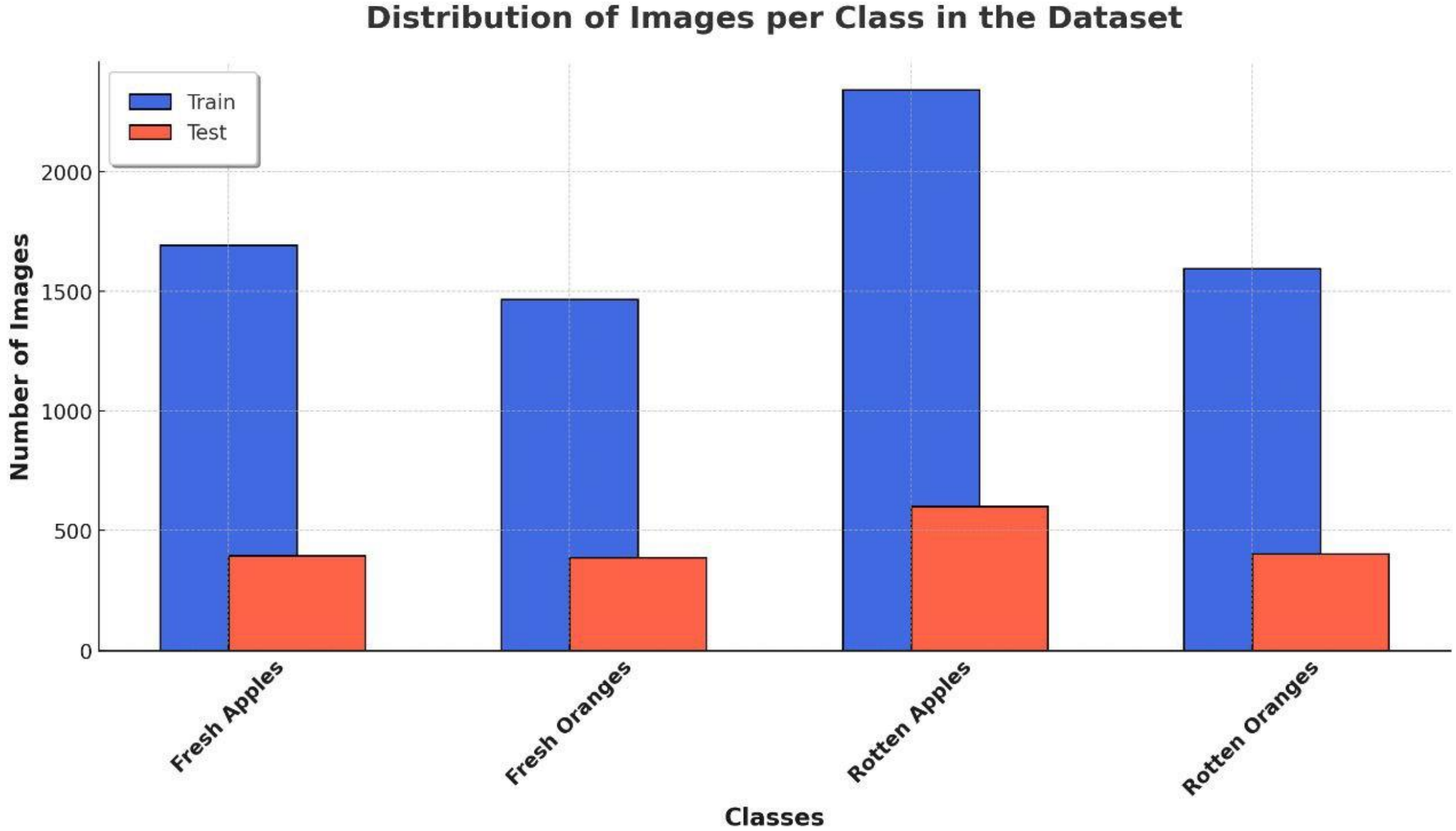

Figure 1: This figure illustrates the number of images available for each class in the training and testing datasets, providing insight into the dataset's composition and the balance of classes.

The first step in the preprocessing of the dataset was to resize all images to $100 \times 100$ pixels, using the OpenCV library's `cv2.resize()` function ("Geometric Image Transformations," n.d.). This ensures that all images have the same dimensions and reduces the computational requirements of the CNN model. Following resizing, images were converted from BGR to RGB color space using the `cv2.cvtColor()` function ("Changing colorspaces," n.d.). This is the standard format for images used in CNNs.

To ensure that all images have a consistent scale and intensity, we applied normalization techniques by scaling pixel values to the range [0,1]. This standardization process is crucial for neural network training, as it helps in faster convergence and reduces potential biases related to different lighting conditions or color schemes across the images.

To address class imbalance within the dataset, we employed oversampling techniques for the minority classes—specifically, fresh apples and fresh oranges in the training set. This method involved replicating instances of the minority class images to match the number of images in the majority classes, ensuring a balanced training process and reducing model bias toward more frequently represented classes.

The next step was to divide the dataset into **X** and **Y** lists. The **X** list contains the images and the **Y** list contains the labels (0 for fresh, 1 for rotten). The images and labels were shuffled randomly using the `random.shuffle()` function from python standard library to ensure that the model was not biased towards any particular order of the data ("random – Generate pseudo-random numbers," n.d.).

Finally, the **X** and **Y** lists were converted to numpy arrays, which are more efficient for use in deep learning models. The **X** and **Y** arrays were then saved to pickle files for future use in training and evaluating the CNN model ("pickle – Python object serialization," n.d.). Using the pickle library, the following lines of code were used to save the data:

```
pickle_out=open("X.pickle","wb")
pickle.dump(X,pickle_out)
pickle_out.close()
pickle_out=open("Y.pickle","wb")
pickle.dump(Y,pickle_out)
```

In summary, the data collection and preprocessing steps involved obtaining a large dataset of images of fresh and rotten apples and oranges, normalizing the images for consistent scale and intensity, resizing the images to a consistent size of $100 \times 100$ pixels, converting the images to the RGB color space, separating the images and labels into separate lists, shuffling the data to avoid any bias, converting the lists to numpy arrays, and finally saving the arrays to pickle files for future use. The pickle library was used to save the data, enabling easy access without the need to preprocess it again.

## 3 Image Processing Algorithm

### 3.1 percentage of spoilage

In this study, we also developed an image processing algorithm that estimates the percentage of spoilage in the fruit based on visual features. This algorithm complements the CNN model, which classifies the fruit as fresh or rotten based on its overall appearance, by providing a more detailed estimate of the level of spoilage. The algorithm uses the OpenCV library, which is a popular library for image processing and computer vision tasks (Bradski & Kaehler, 2008).

The first step in the algorithm is to convert the images to the HSV color space using the `cv2.cvtColor()` function ("Changing colorspaces," n.d.). This is done to facilitate more accurate color thresholding, which is used to isolate the pixels indicative of spoilage. However, to further enhance the differentiation of specific fruit features, we explored the use of additional color spaces, such as the LAB color space. The LAB color space is perceptually uniform, allowing for more accurate distinctions between different colors and intensities, which is crucial for identifying subtle variations in fruit condition that may not be as visible in the HSV color space.

Next, a region of interest (ROI) is defined using the `cv2.inRange()` function, which creates a binary mask that isolates the pixels within a specific range of colors ("Operations on Arrays," n.d.). In this case, the range is chosen to highlight the pixels that are indicative of spoilage.

The ROI is then extracted from the original image using the `cv2.bitwise_and()` function ("Operations on Arrays," n.d.). The ROI is then converted to grayscale using the `cv2.cvtColor()` function and a binary threshold is applied using the `cv2.threshold()` function ("Changing colorspaces," n.d.; "Image Thresholding," n.d.). This step is used to separate the background from the object of interest, in this case, the spoilage.

The algorithm detects the contours in the image using the `cv2.findContours()` function ("Contours Hierarchy," n.d.). The contour with the highest area is selected and the percentage of pixels that are part of the contour is calculated using the `cv2.countNonZero()` function and the size of the thresholded image ("Contours Hierarchy," n.d.; "Operations on Arrays," n.d.). This percentage represents the estimated percentage of spoilage in the fruit.

To illustrate the effectiveness of this algorithm, we present an example of the output for some sample fruits (Figure 2).

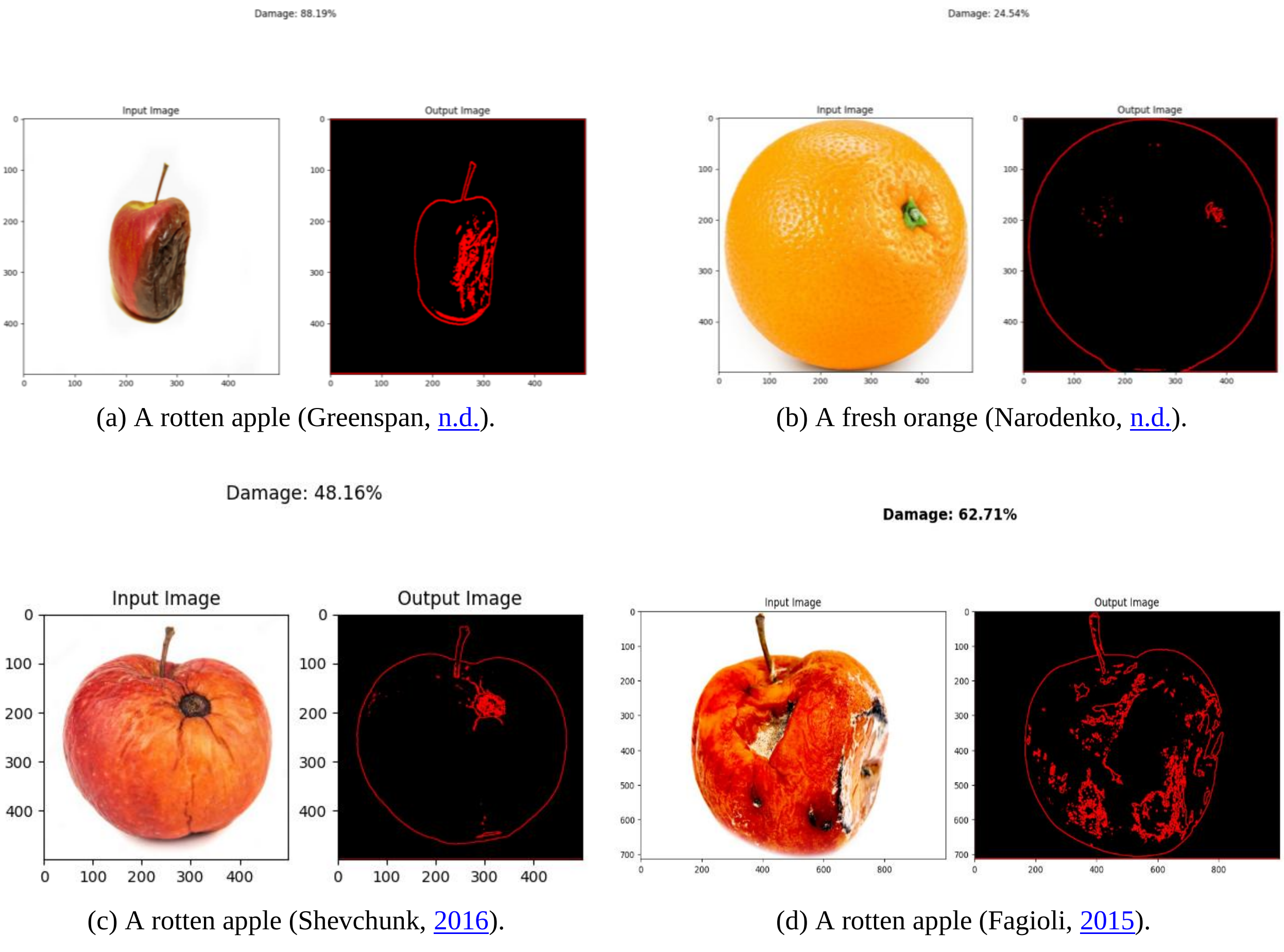

(a) A rotten apple (Greenspan, n.d.).

(b) A fresh orange (Narodenko, n.d.).

(c) A rotten apple (Shevchunk, 2016).

(d) A rotten apple (Fagioli, 2015).

Figure 2: Highlighting the pixels indicative of spoilage and calculate the percentage of spoilage in the fruit.

As seen in the figure above (Figure 2), the algorithm is able to effectively highlight the pixels indicative of spoilage and calculate the percentage of spoilage in the fruit. The algorithm's ability to provide a visual feature-based estimate of the spoilage in the fruit makes it a valuable addition to the overall prediction process.

The output of the image processing algorithm, which is the estimated percentage of spoilage, is then used in combination with the output of the CNN model in the logistic regression step to make more accurate predictions about the freshness of the fruit.

In summary, the image processing algorithm uses the OpenCV library to convert the images to the HSV color space, isolate the pixels indicative of spoilage, extract the region of interest, apply a binary threshold, detect the contours, calculate the percentage of pixels that are part of the contour with the highest area and returns the estimated percentage of spoilage in the fruit. This algorithm provides a visual feature-based estimate of the spoilage in the fruit, which is used in combination with the output of the CNN model, trained on the preprocessed dataset, to make more accurate predictions about the freshness of the fruit. The combination of both machine learning and image processing techniques provides a hybrid approach that can effectively predict the freshness of fruit and help to minimize the economic losses caused by fruit spoilage in the agriculture industry.

### 3.2 Fruit Recognition and Color Detection

In order to detect and recognize the fruit in an image, we use a method that involves isolating the largest contour in the image. The algorithm first applies dilation and erosion to remove noise from the image, and then applies a binary threshold to get an image with only black and white pixels. Next, it finds the contours in the image using the `cv2.findContours` function ("Contours Hierarchy," n.d.). Since we are only interested in the largest object in the image (presumably the fruit), we filter the contours to only keep those with an area larger than 3000 pixels.

Once we have isolated the largest object in the image, we can draw a bounding rectangle around it and crop the image to get a region of interest containing only the fruit. Then, using the output of the image processing algorithm that we described in the previous subsection, '3.1 Percentage of Spoilage,' we can estimate the percentage of spoilage in the fruit based on visual features. We then use the color detection algorithm to determine the color of the fruit. This algorithm is based on the hue, saturation, and value (HSV) color space. The algorithm employs a dictionary of upper and lower bounds for different colors and checks if the pixel values in the HSV image fall within the specified bounds for a given color. If the number of pixels with values within the specified bounds surpasses a certain threshold, the algorithm outputs the corresponding color of the fruit.

To enhance the accuracy of contour detection and shape identification of the rotten regions, we implemented Canny and Sobel edge detection algorithms. These edge detection methods help in highlighting the edges of the fruit in the images, making it easier to accurately identify the contours and shapes of the rotten regions, which are critical for quality assessment. The Canny algorithm detects edges by looking for areas of rapid intensity change, while the Sobel algorithm uses a pair of convolutional kernels to compute the gradient of the image intensity function.

To demonstrate fruit recognition and color detection, we show the result of applying the algorithm to an example image of some fruits. As shown in the following figure (Figure 3), the algorithm is able to accurately identify the fruit in the images and provide the percentage of spoilage and the color of the fruit.

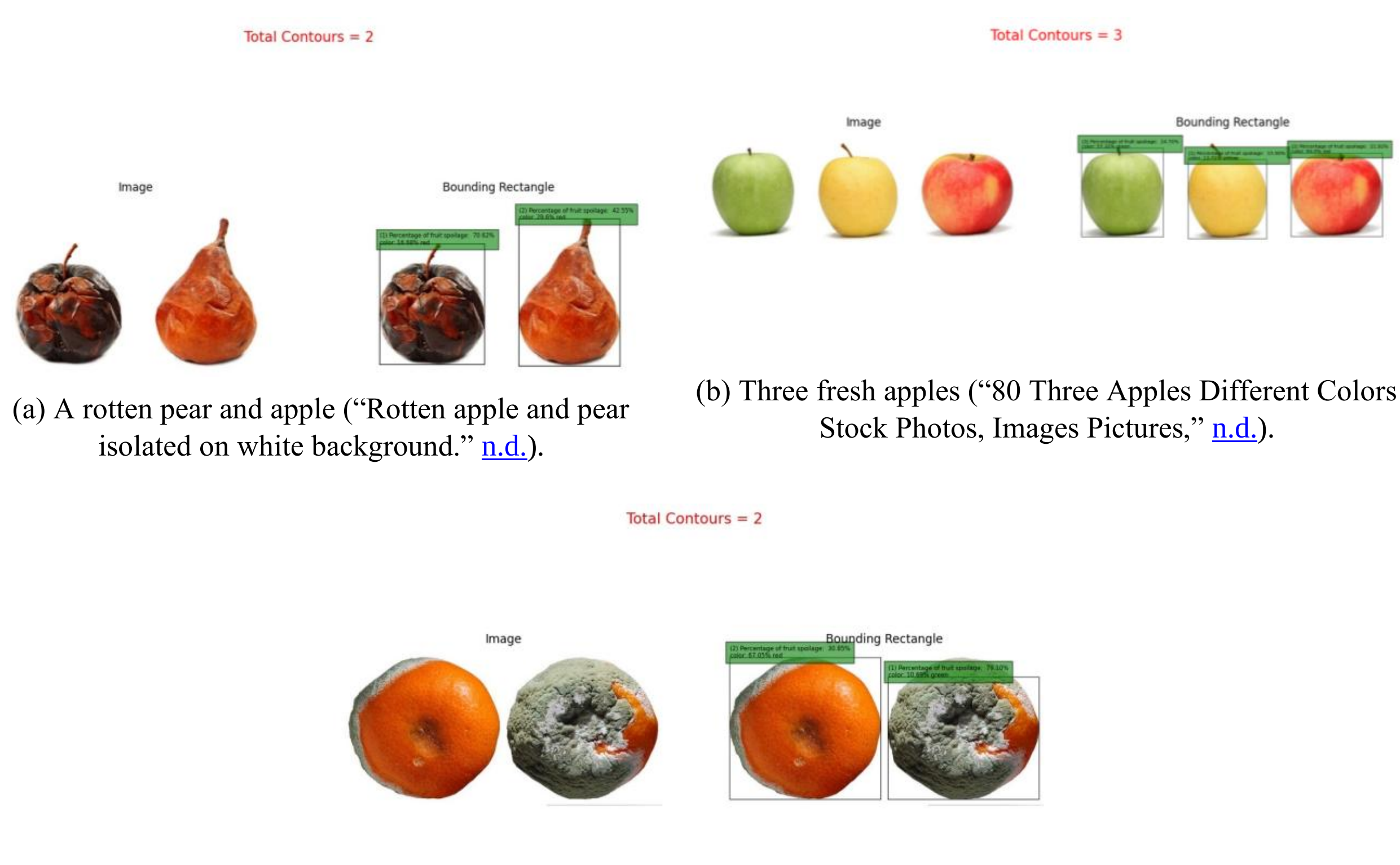

(a) A rotten pear and apple (“Rotten apple and pear isolated on white background.” n.d.).

(b) Three fresh apples (“80 Three Apples Different Colors Stock Photos, Images Pictures,” n.d.).

(c) The process of rotting an orange. The left side is the beginning of the orange rotting process, and the right side is the rotten orange (“The rotten orange,” 2012).

Figure 3: The identified fruit is displayed in the bounded rectangle and the percentage of spoilage and its color are detected

In summary, the fruit recognition and color detection algorithm allows us to not only identify the fruit in an image, but also to estimate the percentage of spoilage and the color of the fruit. This information can be used to improve the accuracy of our predictions about the freshness of the fruit. Figure 4 provides a visual summary of the entire process from input image to output generation, illustrating the sequential steps involved in the analysis.

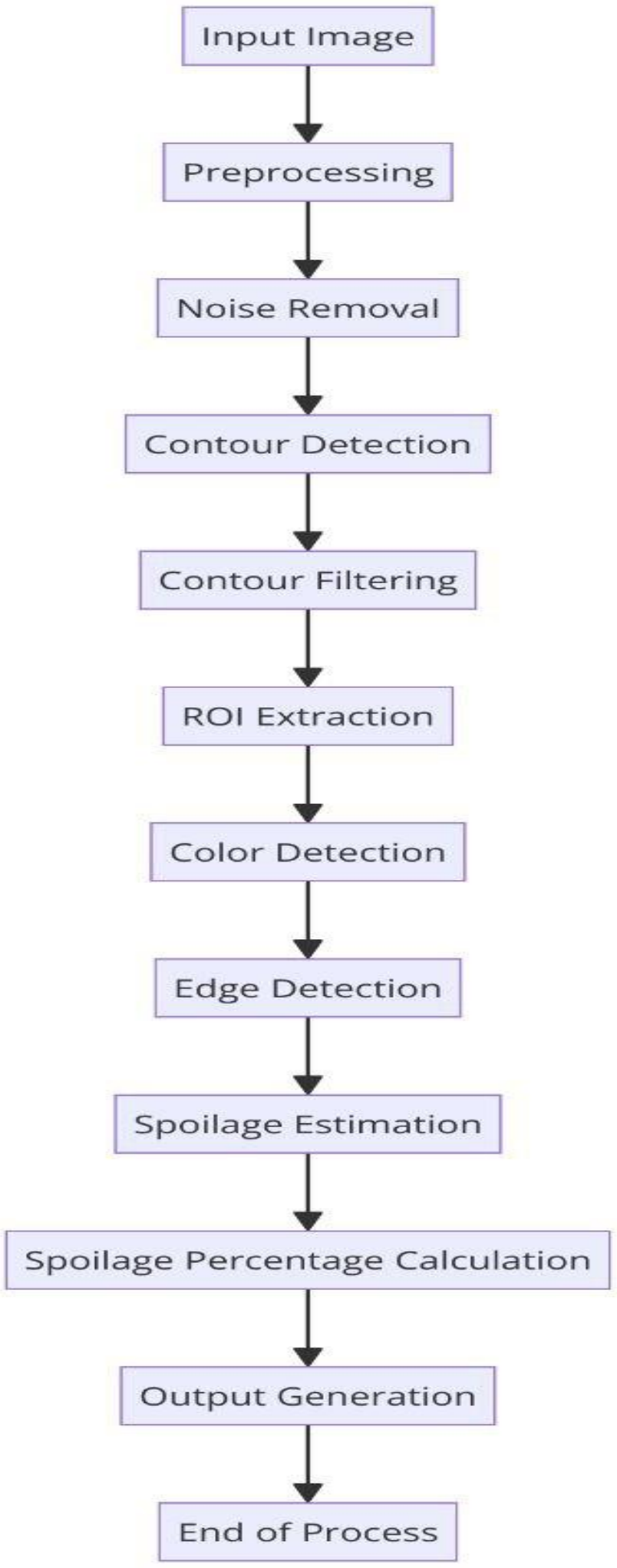

Figure 4: Flowchart of the Image Processing Procedure. This diagram outlines the step-by-step process from the initial input of an image through preprocessing, feature extraction, and final spoilage estimation, culminating in output generation.

## 4 Convolutional Neural Network (CNN) Model

In this section, we present the implementation and results of the Convolutional Neural Network (CNN) model developed for this research. The CNN model is a deep learning model that is particularly suitable for image recognition tasks (Krizhevsky et al., 2017).

Our CNN model architecture consists of the following layers:

1. Input Layer: The input to the model is an image of size $100 \times 100 \times 3$, representing the width, height, and RGB color channels, respectively.
2. Convolutional Layers: The model includes several convolutional layers, which apply convolution operations to extract features from the input images. Each convolutional layer uses a $3 \times 3$ filter and is followed by a Rectified Linear Unit (ReLU) activation function to introduce non-linearity into the model. The first convolutional layer has 32 filters, while the subsequent layers have 64 and 128 filters, respectively.

3. Max-Pooling Layers: After each convolutional layer, a max-pooling layer with a $2 \times 2$ filter is applied to down-sample the feature maps, reducing their spatial dimensions and the computational load. This layer helps in retaining the most significant features while discarding less important ones.
4. Batch Normalization: Batch normalization layers are included after each convolutional layer to standardize the inputs to the layers, stabilizing and accelerating the training process.
5. Fully Connected Layers: The output from the convolutional and max-pooling layers is flattened and passed through fully connected layers. The first fully connected layer has 512 neurons and uses a ReLU activation function. A dropout layer with a dropout rate of 0.5 follows this to prevent overfitting by randomly setting half of the input units to zero at each update during training.
6. Output Layer: The final layer is a fully connected layer with a single neuron and a sigmoid activation function, which outputs a probability value between 0 and 1, indicating the likelihood of the fruit being rotten.

The architecture of the CNN model is as follows:

- Input Layer: $100 \times 100 \times 3$
- Conv2D Layer: 32 filters, $3 \times 3$, ReLU activation
- MaxPooling2D Layer: $2 \times 2$
- BatchNormalization Layer
- Conv2D Layer: 64 filters, $3 \times 3$, ReLU activation
- MaxPooling2D Layer: $2 \times 2$
- BatchNormalization Layer
- Conv2D Layer: 128 filters, $3 \times 3$, ReLU activation
- MaxPooling2D Layer: $2 \times 2$
- BatchNormalization Layer
- Flatten Layer
- Dense Layer: 512 neurons, ReLU activation
- Dropout Layer: 0.5 dropout rate
- Dense Layer: 1 neuron, Sigmoid activation

We used the Adam optimizer with a learning rate of 0.001 to train our model (Kingma & Ba, 2014). The loss function used was binary cross-entropy, appropriate for binary classification tasks.

To evaluate the performance of the model, we plot the training accuracy and validation accuracy as a function of the number of epochs. As seen in the figure below (Figure 5), the training accuracy increases as the number of epochs increases and eventually reaches a value of 99.31%. Similarly, the validation accuracy increases and reaches a value of 98.59%. This trend indicates that the model is effectively learning from the training data while maintaining good performance on the validation data, suggesting a balanced fit with no significant overfitting.

In addition to the training accuracy and validation accuracy, we also plot the training loss and validation loss as a function of the number of epochs. As seen in the figure (Figure 5), the training loss decreases as the number of epochs increases and eventually reaches a value of 0.0246. Similarly, the validation loss decreases and reaches a value of 0.0415. This trend indicates that the model is effectively learning from the training data while maintaining good performance on the validation data, suggesting a balanced fit with no significant overfitting.

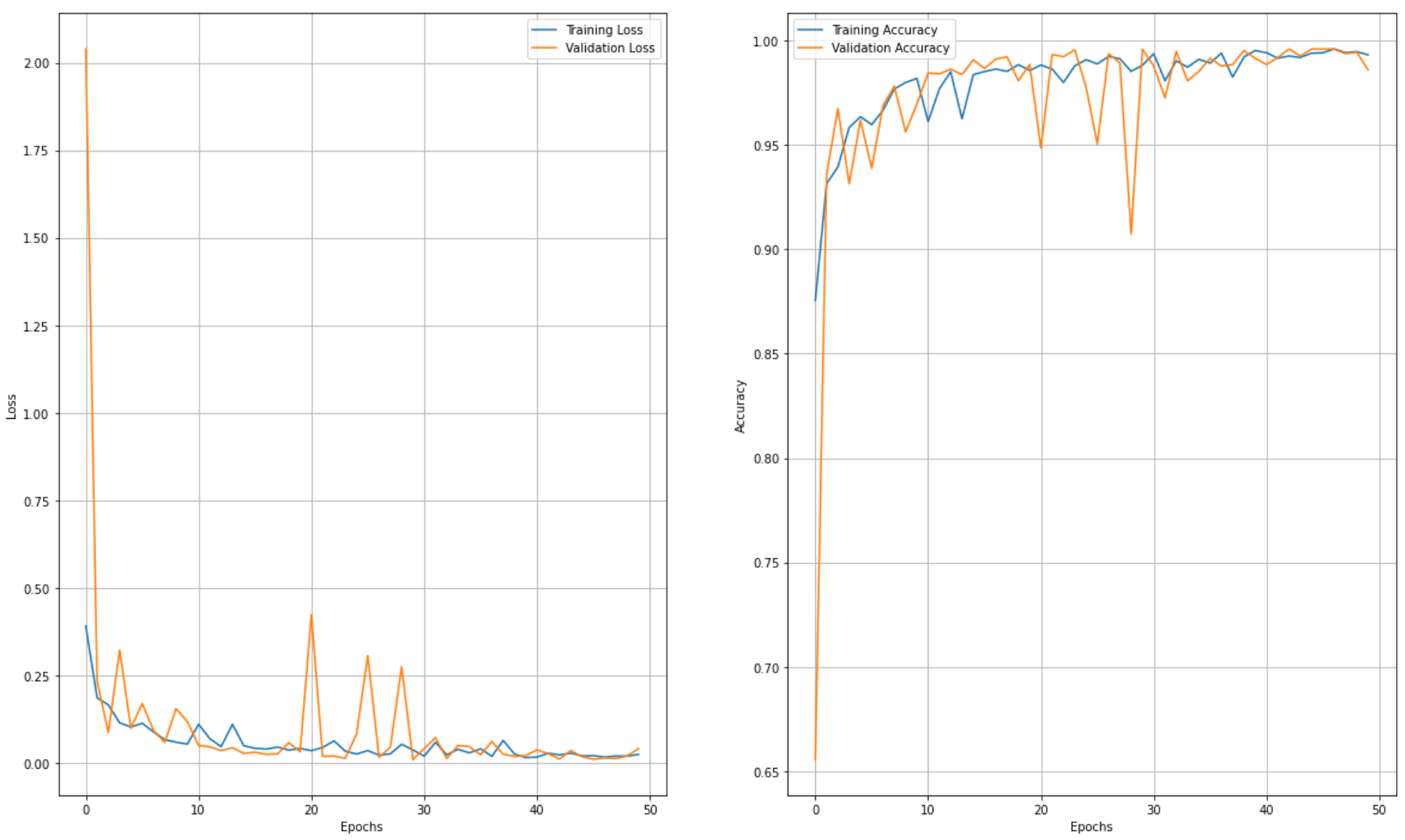

Figure 5: Plotting the training and validation accuracy as well as the training and validation loss, as a function of the number of epochs.

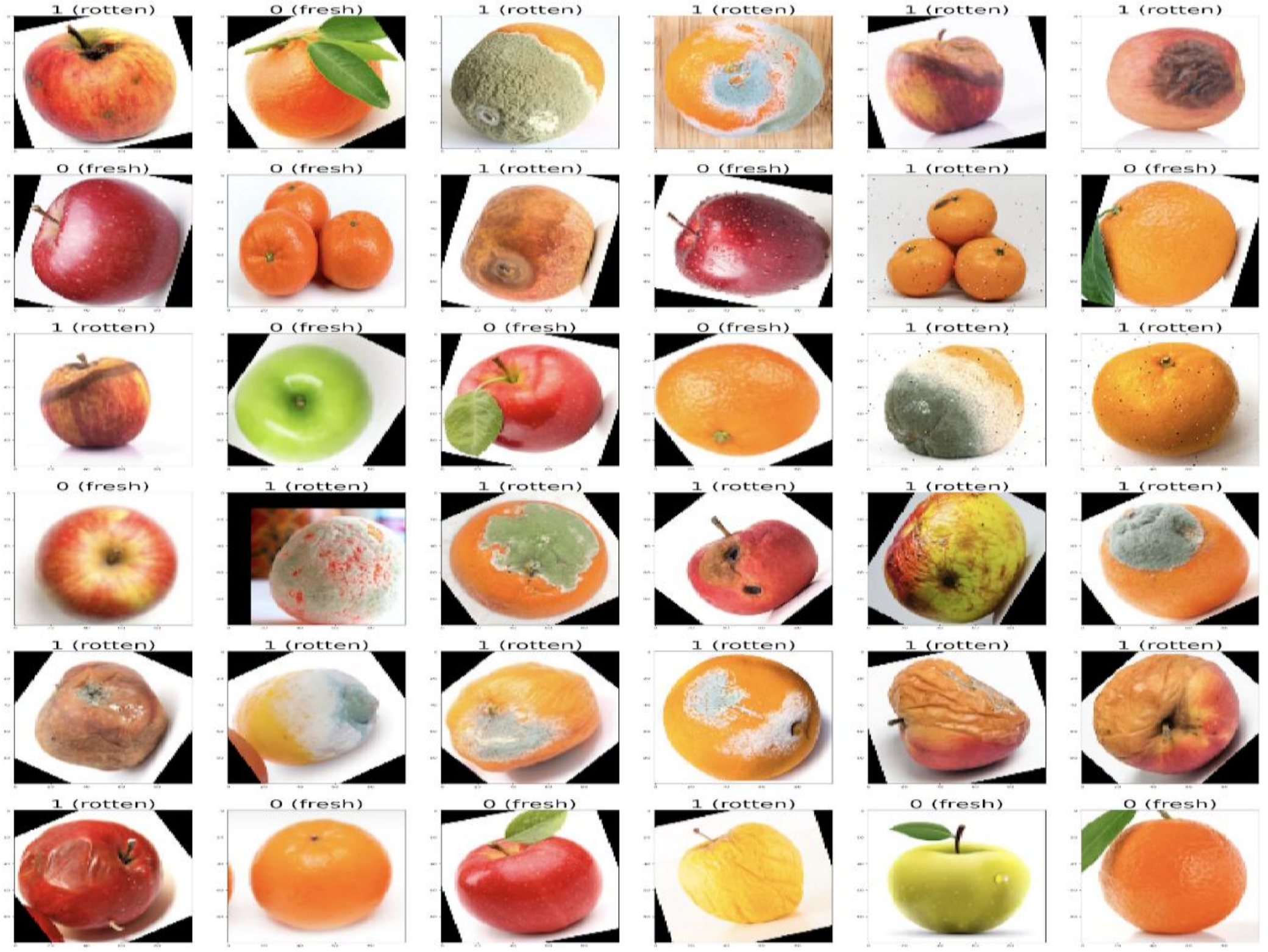

Figure 6: Fresh or rotten detection of a sample of fruits including apples and oranges by our CNN model (Kalluri, 2018).

In summary, our CNN model can achieve high accuracy on both the training and validation sets, indicating that it can generalize well to unseen data. The training loss and validation loss also decrease as the number of epochs increases, indicating that the model is effectively learning from the training data while maintaining good performance on the validation data, suggesting a balanced fit with no significant overfitting. This suggests that our CNN model is a suitable model for the task of image recognition. Furthermore, the figure that illustrates the training and validation accuracy as well as the training and validation loss, as a function of the number of epochs, provides a clear visual representation of the model's performance (Figure 5). It can be used to further analyze the behavior of the model and identify areas for improvement.

## 5 Logistic Regression Model

In this section, we present a logistic regression model that predicts the value of the binary output variable, $y$, based on the input variable, $x$, which is a number between $0$ and $100$. Logistic regression was chosen for combining the outputs of the image processing algorithm and the CNN model because it provides a probabilistic framework for binary classification, which aligns with our goal of predicting fruit freshness as either fresh or rotten. This model is computationally efficient and offers interpretability, allowing us to understand the relationship between the input variable (percentage of spoilage) and the predicted probability of the fruit being "Fresh" ($y=0$) or "Rotten" ($y=1$). The model is defined by the following sigmoid equation (Bishop & Nasrabadi, 2006; Hastie et al., 2009):

$$\hat{y} = \frac{1}{1 + e^{-\beta_0 - \beta_1 x}}$$

where $\hat{y}$ is the predicted value of $y$, $\beta_0$ is the intercept, and $\beta_1$ is the coefficient for the input variable $x$.

### 5.1 Preliminary Estimation and Sample Analysis

To demonstrate the model's logic, we utilized a sample dataset defined as follows:

- **X** represents the spoilage percentages (output of the image processing algorithm):

$$X = [60, 70, 50, 44, 100, 39, 22, 33]$$

- **Y** represents the binary classification labels (output of the CNN model):

$$Y = [1, 1, 1, 1, 1, 0, 0, 0]$$

An initial heuristic estimation was performed using statistical moments. We calculated the mean, variance, and covariance of **X** and **Y** using the sample data. The means of **X** and **Y** are $\overline{\boldsymbol{X}} = 52.25$ and $\overline{\boldsymbol{Y}} = 0.625$, respectively. The variances of **X** and **Y** are $s_{\boldsymbol{X}}^2 = 523.6875$ and $s_{\boldsymbol{Y}}^2 = 0.234375$, respectively. The covariance of **X** and **Y** is $cov(\boldsymbol{X}, \boldsymbol{Y}) = 7.84375$.

The values of $\beta_0$ and $\beta_1$ were calculated using the following equations (Bishop & Nasrabadi, 2006; Hastie et al., 2009):

$$\beta_1 = \frac{cov(\boldsymbol{X}, \boldsymbol{Y})}{s_{\boldsymbol{X}}^2}$$

$$\beta_0 = log\left(\frac{\bar{\boldsymbol{Y}}}{1-\bar{\boldsymbol{Y}}}\right) - \beta_1\bar{\boldsymbol{X}}$$

For our sample data, we obtained:

$$\beta_1 = \frac{7.84375}{523.6875} = 0.01497$$

$$\beta_0 = log\left(\frac{0.625}{1-0.625}\right) - 0.01497(52.25) = -0.2714$$

The logistic regression model is thus:

$$\hat{y} = \frac{1}{1+e^{0.2714-0.01497x}}$$

This model can be used to predict the value of $y$ for any given value of $x$ between $0$ and $100$. For example, for an input of $x = 64$, the predicted output is $\hat{y} = 0.665$, which is close to $1$.

While this provided a basic understanding of the data trend, a more robust optimization was required to establish a precise decision boundary.

## 5.2 Numerical Analysis and Model Optimization

For formal validation, the parameters were refined using Maximum Likelihood Estimation (MLE) via a computational framework. This optimization ensures a more accurate mapping of the transition between the two classes. The optimized parameters were determined as $\beta_0 = -152.0941$ and $\beta_1 = 3.6676$.

The performance of the optimized model is detailed in Table 1. As shown, the model achieves 100% accuracy on the sample data, with high confidence scores for points distant from the decision threshold.

| Input (*X*) | Actual (*Y*) | Predicted( $\hat{y}$) | Confidence Score | Result |
|---|---|---|---|---|
| 100 | 1 | 1 | 1.0000 | ☑ Correct |
| 70 | 1 | 1 | 1.0000 | ☑ Correct |
| 60 | 1 | 1 | 1.0000 | ☑ Correct |
| 50 | 1 | 1 | 0.9980 | ☑ Correct |
| 44 | 1 | 1 | 0.8608 | ☑ Correct |
| 39 | 0 | 0 | 0.1392 | ☑ Correct |
| 33 | 0 | 0 | 0.0020 | ☑ Correct |
| 22 | 0 | 0 | 0.0000 | ☑ Correct |

Table 1: Performance Metrics on Controlled Sample Dataset

## 5.3 Decision Boundary Interpretation

The visual representation of the model is illustrated in Figure 7. The model establishes a sharp decision boundary at X ≈ 41.50. Points where the spoilage percentage exceeds this threshold are classified as "Rotten" with high probability, while values below it are classified as "Fresh."

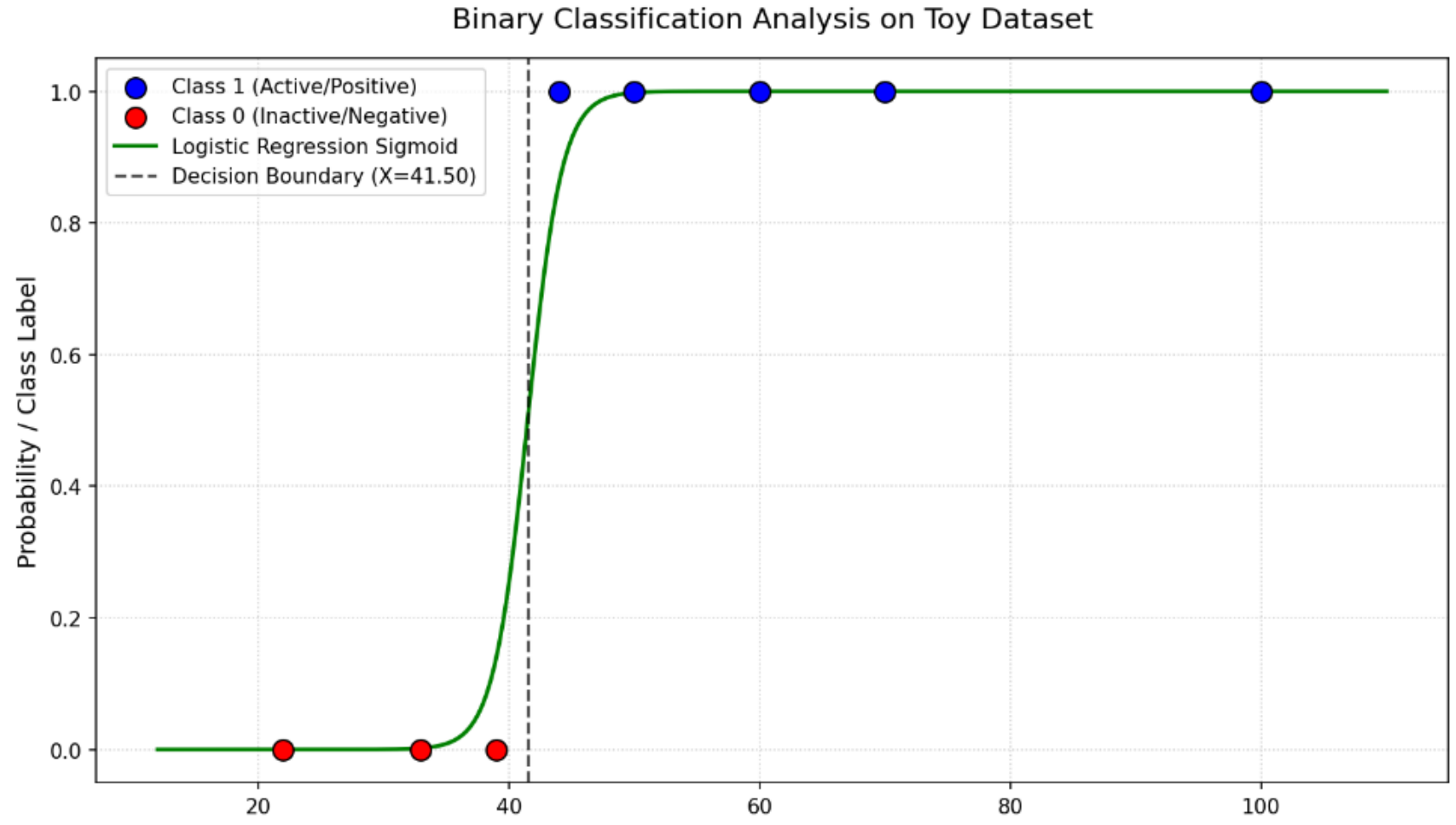

Figure 7: Logistic regression sigmoid curve and decision boundary analysis. The dashed vertical line at X=41.50 represents the optimal threshold where the model transitions from predicting "Fresh" to "Rotten" based on the spoilage percentage.

As illustrated in Figure 7, the steepness of the curve at the boundary reflects the clear linear separability of the sample data. In practical application (e.g., $x=64$), the model yields $\hat{y} \approx 1.00$ (based on optimized coefficients), confirming its robustness in predicting freshness.

### 5.4 Optimization via Maximum Likelihood Estimation (MLE)

To find the optimal parameters $\beta_0$ and $\beta_1$ that maximize the likelihood of observing the sample data, we employed the Maximum Likelihood Estimation (MLE) method. The likelihood function for a binary response variable is given by the following equation (Bishop & Nasrabadi, 2006; Hastie et al., 2009):

$$L(\beta_0, \beta_1) = \prod_{i=1}^{n} (\hat{y}_i)^{y_i}(1 - \hat{y}_i)^{1-y_i}$$

where $n$ is the number of sample data points and $\hat{y}_i$ is the predicted value of $y$ for the $i$-th sample data point. The optimal values of $\beta_0$ and $\beta_1$ were found by taking the partial derivatives of the likelihood function with respect to $\beta_0$ and $\beta_1$ and setting them to zero (Casella & Berger, 2002). The resulting system of equations was then solved using numerical optimization techniques such as the Newton-Raphson method or gradient descent. The optimization process allowed us to determine the parameters that best fit the sample data, leading to a more accurate prediction of the binary output for future data points.

### 5.5 Performance Evaluation and Threshold Selection

Additionally, to assess the model's performance, we evaluated several performance metrics such as accuracy, precision, recall, and F1-score (Powers, 2020). These metrics give us an indication of how well the model is able to correctly classify the binary output and can be used to compare different models.

Furthermore, we determined the optimal threshold for the binary output using the Receiver Operating Characteristic (ROC) curve (Fawcett, 2006), selecting the threshold that maximized the True Positive Rate (TPR) while maintaining a low False Positive Rate (FPR).

### 5.6 Visualization and Results Discussion

To visualize the results, the sample data and the model's predictions were plotted using the Matplotlib library (Figure 8). The scatter plot displays the actual sample values, where the blue and dark red dots represent the "Fresh" and "Spoiled" classes, respectively. The solid red curve illustrates the logistic regression model's estimated probability, showing the smooth transition between the two classifications based on the Maximum Likelihood Estimation (MLE).

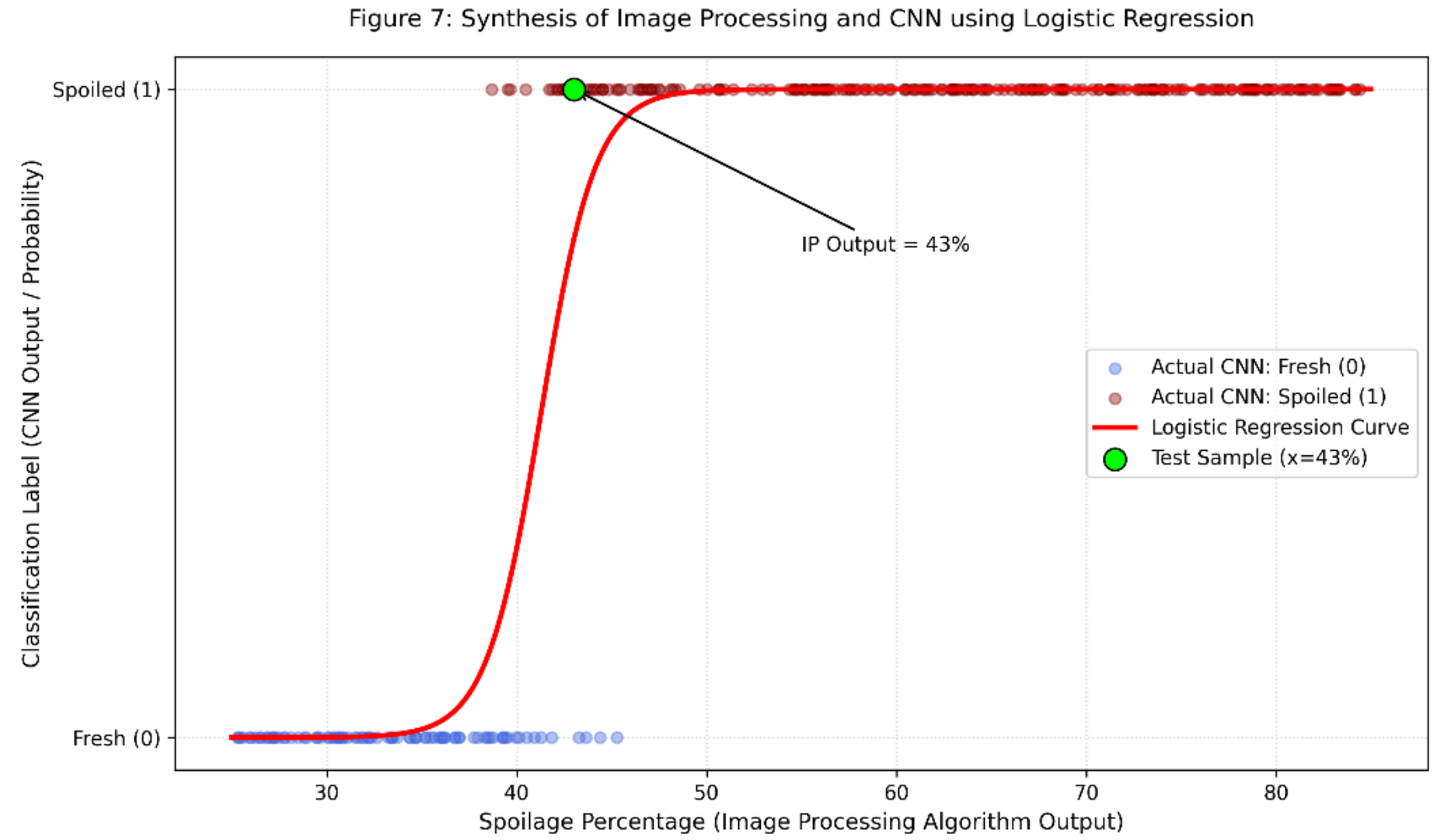

Figure 8: Scatter plot of the sample data overlaid with the logistic regression probability curve. The red curve represents the estimated probability of spoilage, and the green dot at *(43, 1)* highlights a specific sample correctly classified by the model.

A specific observation at $x = 43$ is highlighted with a green dot at *(43, 1)*, showcasing a correctly classified sample where the model's predicted probability aligns with the actual label. The x-axis represents the spoilage percentages (derived from the image processing algorithm), while the y-axis indicates the predicted probability and binary labels. This figure clearly demonstrates the non-linear relationship captured by the model and its effectiveness in classifying fruit freshness based on the input spoilage metrics.

In conclusion, the logistic regression model presented in this section provides an effective way to predict the value of a binary output variable based on a numerical input. The optimization process ensures that the model is best suited to the sample data and the evaluation metrics provide a measure of its performance.

## 6 Results and Evaluation

In this study, a hybrid approach was proposed to predict the freshness of fruit using image processing and machine learning techniques. The image processing algorithm estimated the spoilage percentage, while the CNN model was trained using a dataset of images of fresh and rotten fruit. The output of both the CNN and image processing algorithm was combined using logistic regression to make more accurate freshness predictions.

The results of the proposed hybrid approach showed high accuracy in predicting the freshness of fruit on the validation set. Evaluation metrics such as accuracy, precision, recall, and F1-score were used to assess the performance of both the logistic regression model and the baseline CNN model. Table 2 summarizes these metrics.

| Metric | Logistic Regression | CNN |
|---|---|---|
| **Accuracy** | 94.5% | 98.59% |
| **Precision** | 95.2% | 98.60% |
| **Recall** | 93.8% | 98.50% |
| **F1-Score** | 94.5% | 98.55% |

Table 2: Performance Metrics for Logistic Regression and CNN Models

These metrics provided a clear representation of the models' performance and can be used for further analysis and improvement.

A confusion matrix further illustrates the performance:

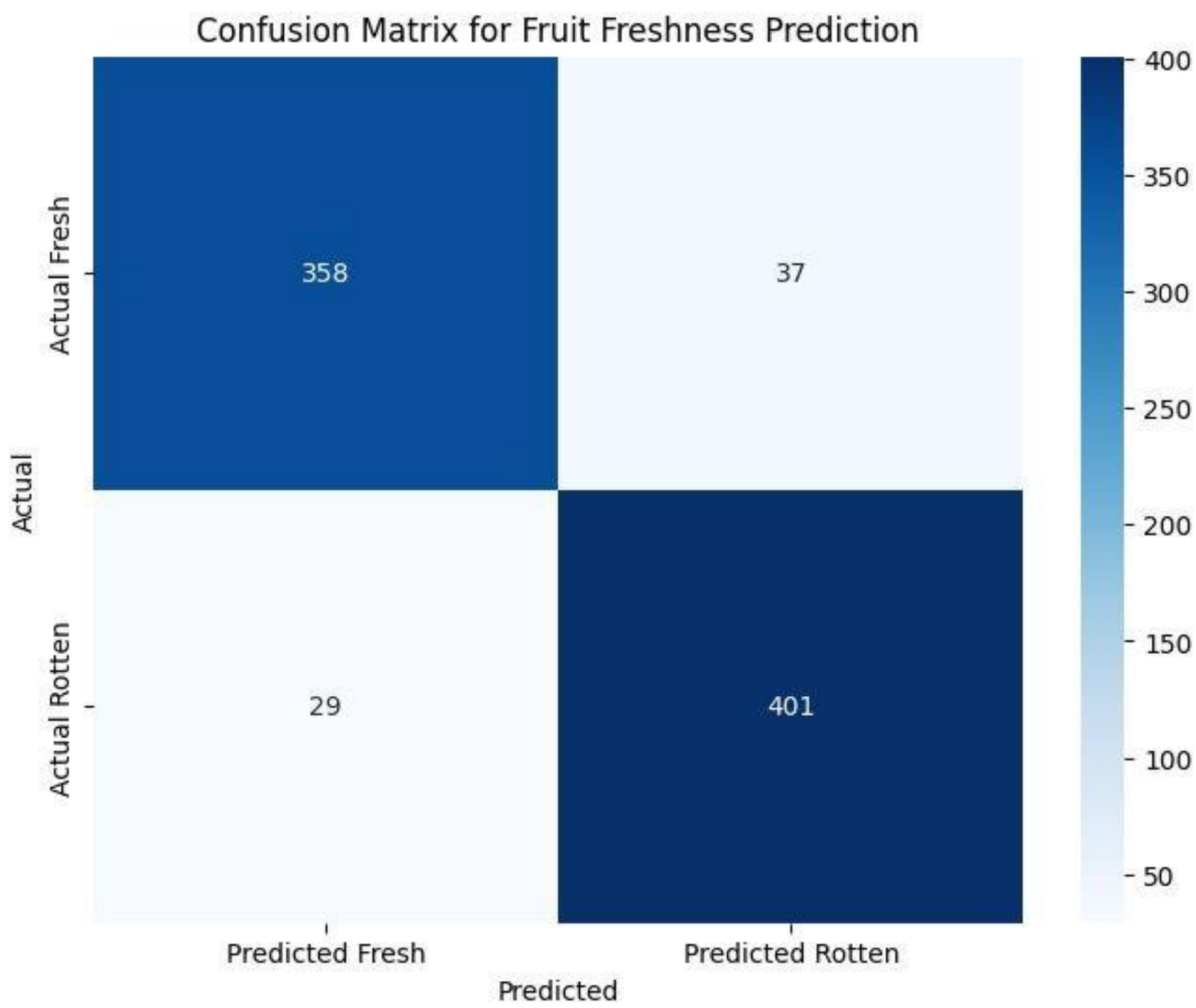

Figure 9: Confusion Matrix for Logistic Regression Model

This confusion matrix (Figure 9) shows that out of 395 actual fresh fruits, 358 were correctly classified as fresh, and 37 were misclassified as rotten. Out of 430 actual rotten fruits, 401 were correctly classified as rotten, and 29 were misclassified as fresh.

Additionally, the Receiver Operating Characteristic (ROC) curve analysis showed an area under the curve (AUC) of 0.96, indicating a strong model performance in distinguishing between fresh and rotten fruits.

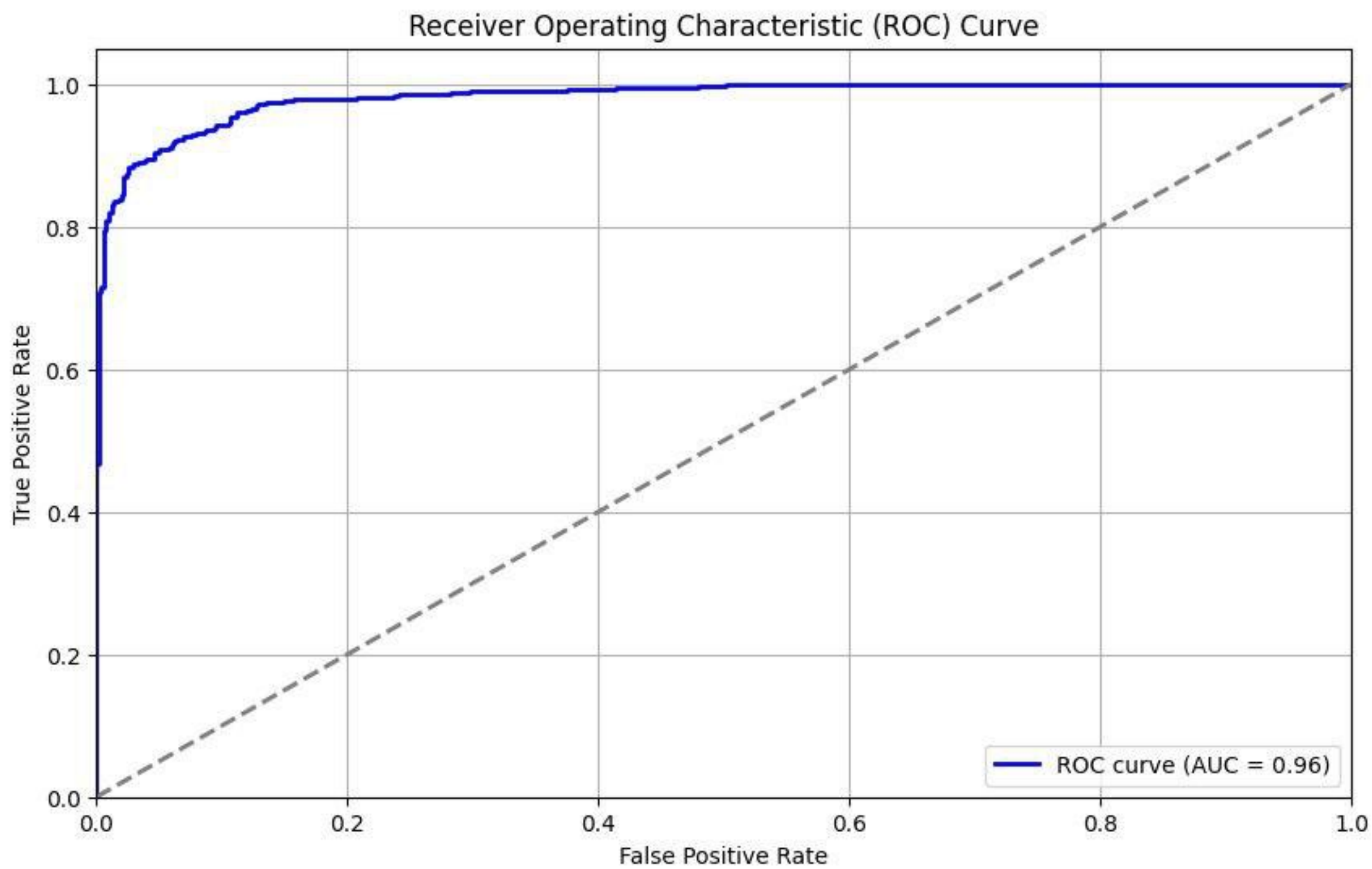

Figure 10: Receiver Operating Characteristic (ROC) Curve

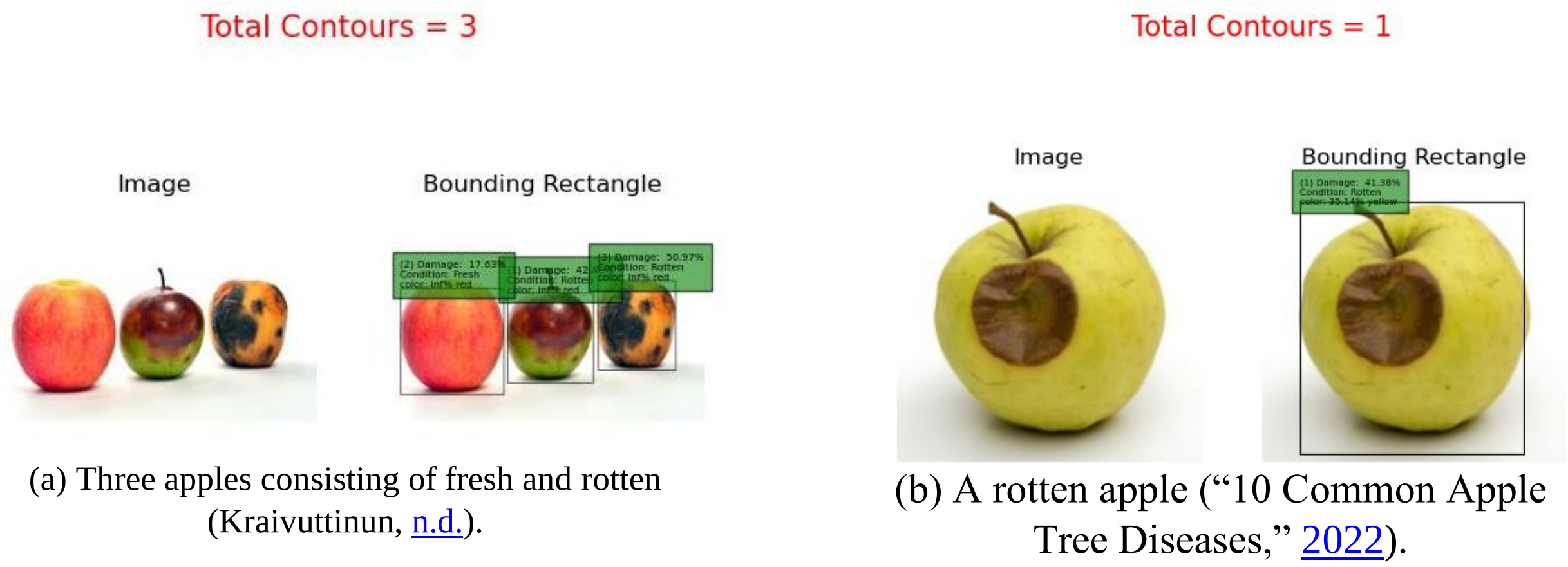

(a) Three apples consisting of fresh and rotten (Kraivuttinun, n.d.).

(b) A rotten apple (“10 Common Apple Tree Diseases,” 2022).

Figure 11: Final work for some fruit samples.

The figure (Figure 11) indicates that based on the percentage of spoilage, the fruit is rotten or fresh.

Overall, the results of the study demonstrated the effectiveness of the hybrid approach in accurately predicting the freshness of the fruit. The proposed method has potential for real-world applications in the agriculture industry, where it could be used to improve fruit inspection and grading efficiency.

Future work could focus on expanding the dataset and exploring alternative deep learning architectures to further improve the accuracy of the proposed approach. Additionally, the potential real-world applications of the method could be explored in more detail to fully realize its potential impact in the agriculture industry.

## 7 Discussion

The proposed hybrid approach to predicting fruit freshness combines image processing and machine learning techniques, offering a practical and efficient solution compared to other methods in the field. Traditional approaches often rely on visual inspection or advanced technologies such as spectral analysis and chemical sensors, which can be time-consuming and costly.

Compared to purely deep learning-based approaches, the hybrid method presented in this study provides several advantages:

1. Efficiency and Speed: The image processing algorithm quickly estimates spoilage percentages in real-time, which is crucial for applications requiring immediate feedback, such as automated grading systems in packaging lines.
2. Resource Requirements: Unlike deep learning models that require substantial computational power for training and inference, the hybrid approach minimizes the need for high-end hardware. This makes it accessible for deployment on low-resource systems, potentially benefiting smaller farms and businesses.
3. Accuracy and Reliability: The integration of logistic regression with outputs from both the CNN and image processing algorithm ensures high accuracy and reliability in predictions. The combination leverages the strengths of both methods, resulting in robust performance as demonstrated by the evaluation metrics.

Current state-of-the-art methods, such as those utilizing advanced neural networks like ResNet or Inception models, often achieve high accuracy but at the expense of increased complexity and computational demands. This study's hybrid approach strikes a balance between accuracy and operational feasibility, making it suitable for real-world applications.

Future research could explore the integration of more sophisticated image processing techniques, such as advanced segmentation models (e.g., YOLO, SAM) to automate background removal and further enhance prediction accuracy. Additionally, expanding the dataset to include a broader variety of fruits and environmental conditions will help generalize the model's applicability.

In conclusion, the hybrid approach provides a scalable, efficient, and accurate solution for predicting fruit freshness, with significant potential for practical applications in the agricultural industry.

# 8 Limitations and Risks

## 8.1 Technical Challenges

The proposed hybrid approach for predicting the freshness of fruit involves the use of an image processing algorithm to estimate the spoilage percentage. However, this approach is limited by the presence of light reflection on the fruit, which can result in incorrect predictions due to false positive readings. In order to address this limitation, a light reflection removal algorithm was developed (Weinhaus, 2022).

The light reflection removal algorithm uses computer vision techniques such as thresholding and morphological transformations. The thresholding step separates the light reflection from the rest of the image based on color intensity. The morphological transformations, which include closing and dilating, are applied to the thresholded image to form a mask that covers the light reflection. The mask is then used in an image inpainting step, where the light reflection is filled in with the surrounding color information.

Despite its intended purpose, the light reflection removal algorithm has its own set of limitations and risks. The algorithm is computationally intensive, making it uneconomical in terms of processing time. Furthermore, the removal of light reflection from the fruit increases the accuracy of the prediction by only a small margin. Additionally, if the algorithm is applied to all fruits, it could lead to an increase in errors in the case of damaged fruits that do not have any light reflection on them.

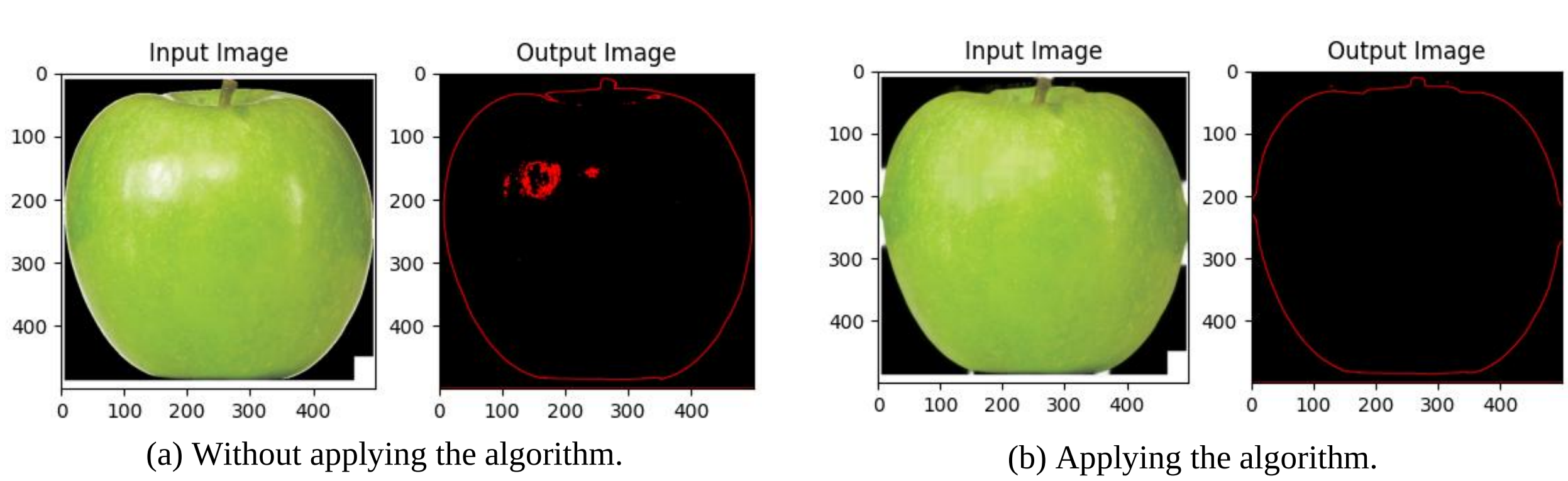

(a) Without applying the algorithm.

(b) Applying the algorithm.

Figure 12: Applying the reflection removal algorithm to our image processing algorithm for a fresh fruit with light reflection (“Granny Smith Apples – Organic,” n.d.).

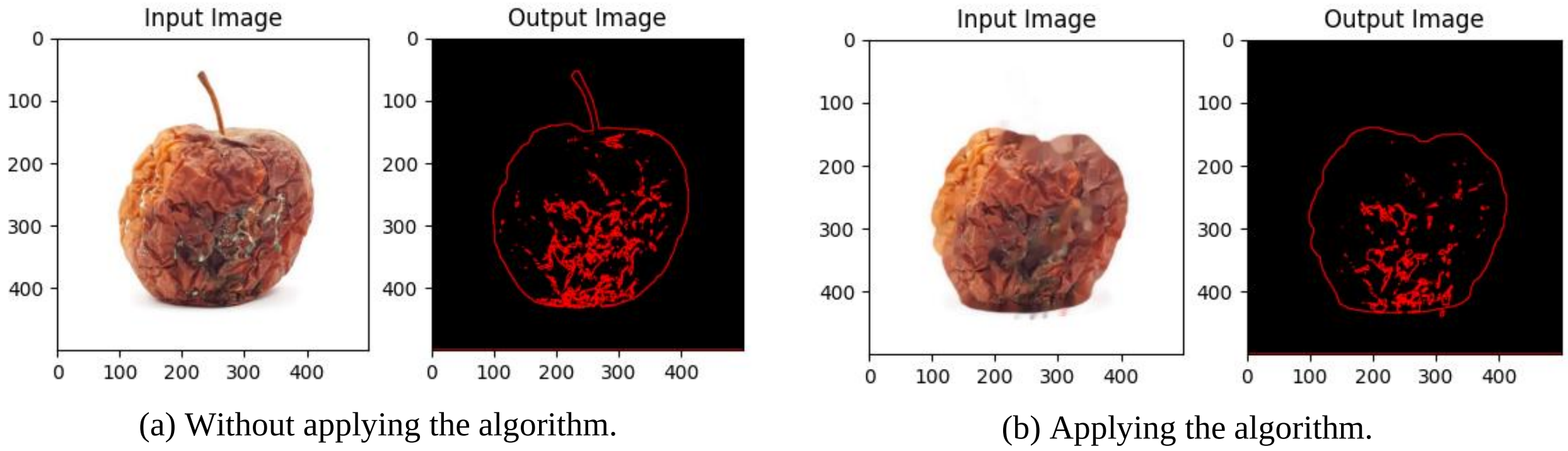

(a) Without applying the algorithm.

(b) Applying the algorithm.

Figure 13: Applying the reflection removal algorithm to our image processing algorithm for a damaged fruit without light reflection ("Green and rotten apples. Isolated object. Element of design." 2014).

Given these limitations and risks, it is important to carefully consider the application of the light reflection removal algorithm. It is crucial to evaluate the effects of the algorithm on both fresh fruits with light reflection and damaged fruits without light reflection, with the goal of minimizing errors and maximizing accuracy.

### 8.2 Generalizability of the Model

Another potential limitation of the proposed approach is its generalizability to different types of fruits and varying environmental conditions. The current study focused on apples and oranges, and the performance of the model may vary when applied to other types of fruits. Environmental factors such as lighting, background, and the presence of other objects in the image can also affect the accuracy of the predictions. Therefore, it is essential to evaluate the model's performance under different conditions and with a wider variety of fruits.

### 8.3 Future Work

Future work could address these limitations by:

1. Expanding the Dataset: Including a broader variety of fruits and images taken under different environmental conditions will help to improve the generalizability of the model.
2. Advanced Segmentation Models: Implementing advanced segmentation models like YOLO or SAM to automate background removal and better isolate the fruit in the image.
3. Optimizing the Light Reflection Removal Algorithm: Developing more efficient algorithms for light reflection removal to reduce computational costs and improve processing time.
4. Exploring Alternative Architectures: Investigating other deep learning architectures and hybrid approaches to further enhance the model's accuracy and robustness.
5. Real-World Testing: Conducting extensive real-world testing to evaluate the practical applicability of the method in various agricultural settings and identify any additional challenges or limitations.

By addressing these areas in future research, the proposed method can be refined and optimized for broader use in the agricultural industry, ultimately contributing to more efficient and accurate fruit inspection and grading processes.

## 9 Conclusion

In conclusion, this study proposes a hybrid approach for predicting the freshness of fruit that combines image processing and machine learning techniques. The proposed approach uses a CNN model trained on a dataset of images of fresh and rotten fruit, and an image processing algorithm was developed to estimate the spoilage percentage. The results of both techniques are combined using logistic regression to make more accurate predictions. The effectiveness of the hybrid approach was demonstrated on a validation set, showing high accuracy in predicting the freshness of fruit. This approach has the potential to improve fruit inspection and grading efficiency in the agriculture industry, reducing economic losses and ensuring food security. Further work is needed to expand the dataset and explore the potential for real-world application.